%% file: main.tex
\documentclass[runningheads]{llncs}

\usepackage[final]{eccv}

\usepackage{eccvabbrv}

\usepackage{graphicx}
\usepackage{booktabs}
\usepackage[T1]{fontenc}
\usepackage[accsupp]{axessibility}  

\usepackage{overpic}
\usepackage{wrapfig}

\input{preamble}

\usepackage[pagebackref,breaklinks,colorlinks,citecolor=eccvblue]{hyperref}

\usepackage{orcidlink}

\begin{document}
\newcommand\blfootnote[1]{%
  \begingroup
  \renewcommand\thefootnote{}\footnote{#1}%
  \addtocounter{footnote}{-1}%
  \endgroup
}

\title{\nameMethod: Text-driven Consistent Geometry Texturing and Material Assignment}

\titlerunning{\nameMethod: Text-driven Consistent Geometry Texturing}

\author{Duygu Ceylan \and Valentin Deschaintre* \and Thibault Groueix* \and Rosalie Martin \and Chun-Hao Huang \and Romain Rouffet \and Vladimir Kim \and Gaëtan Lassagne}

\authorrunning{D.~Ceylan et al.}

\institute{Adobe}

\newcommand{\nameMethod}{{\mbox{MatAtlas}}\xspace}

\maketitle
\input{sec/0_abstract}    
\input{sec/1_intro}
\input{sec/2_related_work}
\input{sec/3_method}
\input{sec/4_evaluation}
\input{sec/5_conc}
\input{sec/6-ackowledgements}

{
    \small
    \bibliographystyle{splncs04}
    \bibliography{main}
}

\end{document}

%% file: preamble.tex
\usepackage[dvipsnames]{xcolor}


%% file: sec/0_abstract.tex
\begin{abstract}
We present \nameMethod, a method for consistent text-guided 3D model texturing. Following recent progress we leverage a large scale text-to-image generation model (e.g., Stable Diffusion) as a prior to texture a 3D model. We carefully design an RGB texturing pipeline that leverages a grid pattern diffusion, driven by depth and edges. By proposing a multi-step texture refinement process, we significantly improve the quality and 3D consistency of the texturing output. 
To further address the problem of baked-in lighting, we move beyond RGB colors and pursue assigning parametric materials to the assets.
Given the high-quality initial RGB texture, we propose a novel material retrieval method capitalized on Large Language Models (LLM), enabling editabiliy and relightability. 
We evaluate our method on a wide variety of geometries and show that our method significantly outperform prior arts. We also analyze the role of each component through a detailed ablation study. Project page: \url{https://duyguceylan.github.io/matatlas/}
\end{abstract}

%% file: sec/1_intro.tex
\section{Introduction}
\label{sec:intro}
\blfootnote{* Denotes equal contribution.}
Geometry texturing is a critical element in the virtual assets design pipeline. Current industry pipelines rely significantly on manually ``painting" existing geometries which is a skill-demanding, cumbersome and time-consuming process especially for assets that are highly detailed. The recent success of large-scale language-guided image generation models~\cite{Dalle2, StableDiffusion} opens up the possibility to ``paint" images by words instead. Hence, an interesting step is to extend the success of such 2D models to 3D, to provide inspirational texturing workflows. Several recent work~\cite{TEXTure, text2tex} attempts to achieve this goal by iteratively generating 2D images from different view points which are then aggregated in the texture space. Creating a seamless and high quality texture, however, is still a challenging task and existing approaches suffer from typical issues such as the ``Janus problem" (\eg both the front and back of a head being textured with faces). Furthermore, existing approaches focus on generating RGB textures with baked lighting. In practice, however, many applications require relightable assets.

\begin{figure}[t]
    \centering
    \vspace{-1.7 em}
    \includegraphics[width=\linewidth, trim={0cm 0cm 0cm 0cm},clip]{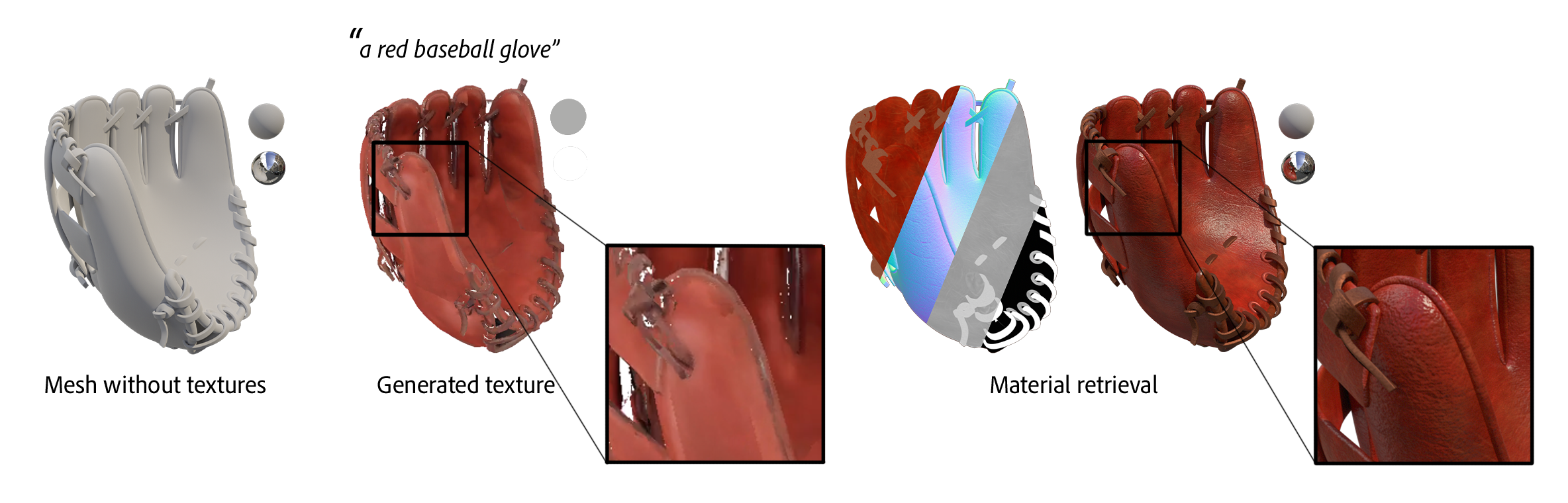}
    \vspace{-1.7 em}
    \captionsetup{type=figure}
        \caption{We present MatAtlas, a novel approach for generation text guided relightable textures for a given 3D mesh. Our method first generates an RGB texture using pre-trained large scale image generation models. Using this texture as a guidance, we retrieve materials from a database and assign to different parts of the mesh resulting in a plausible and fully relightable asset.}
    \label{fig:teaser}
     \vspace{+1.3 em}
\end{figure}%

In this work, we propose a method for texturing 3D assets and using the generated textures to retrieve and assign materials from an existing database to different parts of the asset. In addition to producing relightable assets, we also enable further appearance editing by retrieving parametric materials. In particular, our texturing step follows a similar trend of utilizing a large scale text conditioned image generation model but significantly improves the quality and consistency of the results. This is critical for performing a robust material search where consistent visual cues are necessary to assign coherent materials to different parts of an asset. 
We further observe that humans have strong prior and expectation about what type of materials are suitable for different parts of an asset. We therefore propose a material retrieval algorithm which combines visual cues with global context and priors captured via large language models~\cite{chatgpt}.

Our texturing method uses a \emph{cross-frame attention} strategy to improve the consistency of images generated from different viewpoints. In particular, we generate multiple views in a single image generation pass (we use Stable Diffusion~\cite{StableDiffusion} in our experiments) in a 4x4 grid pattern where each view in the grid can potentially attend to the features of other views in the grid. Compared to the standard view-by-view pipeline \cite{TEXTure,text2tex,texfusion} which processes one view at a time and enforces new view to follow the existing texture, our grid-based approach is more ``synchronous" where all views participate in the generation of each other, improving view consistency.
We further enforce consistency with additional 3D guidance channels through ControlNet~\cite{ControlNet}. Specifically, we use depth conditioning to generate images that are structurally aligned with the 3D asset. We also render occluding and suggestive contours to provide lineart conditioning to avoid texture bleeding across different parts and capture the details of small parts. 
Our texturing method operates in multiple passes to refine the quality of the generated textures. While an initial pass of grid based generation ensures a globally consistent but potentially blurry texture, a second pass aims to refine this texture. We perform a final pass of texture inpainting to guarantee a complete coverage through an improved camera sampling process. 
We evaluate the influence of each of these components in Sec.~\ref{sec:ablation}, showing that their combination significantly improves the final texturing quality.

Given a textured asset and access to a database of material definitions, we propose a novel material retrieval and assignment pipeline. Given a 3D asset with sets of parts, we perform part-based material retrieval. 
Our novel material search algorithm considers both visual cues from the generated texture as well as global context information. For the latter, we exploit the recent success of language models in acquiring priors from large scale data.

Given a 3D asset and a target text prompt, our approach generates high quality and relightable appearance. We compare our method to recent state-of-the-art generative texturing approaches and show superior performance. Our main contributions are as follows:
\begin{itemize}
    \item A complete pipeline that can generate appearances for 3D assets that are relightable and editable.
    \item A generative texturing method that uses large scale text-to-image models that achieves superior performance compared to state of the art.
    \item A novel material retrieval algorithm that robustly matches the generated textures to parametric materials in a database.
\end{itemize}

%% file: sec/2_related_work.tex
\section{Related Work}
\label{sec:related_work}

We briefly review early texturing approaches, then discuss the use of recent vision models for generative texturing.

\vspace{3mm}
\noindent\textbf{Early approaches to texturing.}
Early works~\cite{solidtexturesynthesis, appearancespace, texsynthesisonsurfaces, arbitrarymanifold, Chen:2012:NTT, lu2007context, geomcorrelation, 537718} synthesize 3D textures from an exemplar image by building on classical image synthesis works~\cite{efros2001image,efros1999texture,kwatra2003graphcut,dong2007optimized, kwatra2005texture, portilla2000parametric, wei2009state} using hand-crafted 3D geometric priors and image priors. These methods establish global coherence of the 3D texture but are limited to simple patterns and the texture details usually do not match closely the geometry and its semantic parts.
To address this limitation some techniques proposed to transfer textures from images of similar objects~\cite{Kraevoy2003,Kholgade2014,Tzur_2009_FlexiStickers}, however, they heavily rely on human annotation. One can automate the process by using symmetry, automatic alignment, and co-analysis of collections of assets from the same category~\cite{Wang:2016,photoshape2018}, but these methods cannot be applied to out-of-sample one-of-a-kind shapes and require image references. In contrast our method only needs off-the-shelf pretrained networks. 

\vspace{3mm}
\noindent\textbf{Category-specific models.}
Several works have explored using learning-based methods to generate shape-specific textures~\cite{oechsle2019texture, yu2023texture, siddiqui2022texturify, Berkiten17}. Texture-Fields~\cite{oechsle2019texture} use a variational autoencoder to train a 3D coordinate-based MLP for textures, Point-UV diffusion~\cite{yu2023texture} performs diffusion both on a point cloud and in the UV-space, and Texturify~\cite{siddiqui2022texturify} combines differentiable rendering and a GAN approach. These works are category-specific because of their training data. To go beyond single category models, recent approaches leverage the prior of foundational 2D models trained on a massive amount of image data~\cite{StableDiffusion}.

\vspace{3mm}
\noindent\textbf{Texture generation with vision models.}
Several works propose to use CLIP as an energy to optimize mesh properties like texture and geometry~\cite{text2mesh, clipmesh, textdeformer}. Dreamfusion~\cite{dreamfusion} and Score Jacobian Chaining~\cite{sjc} concurrently propose Score-Distillation Sampling (SDS) to distill 2D gradients from a sampling diffusion model, which strictly improves over CLIP~\cite{fantasia3d, magic3d, prolificdreamer, sweetdreamer, dreamgaussian, magic123, textmesh, long2023wonder3d}. Latent-Nerf~\cite{latentnerf} proposes \textit{latent-paint}, a technique to optimize a texture in the latent space of Stable Diffusion.
These approaches suffer from a few drawbacks. First, they require many forward passes through large vision models which makes them relatively slow. Second, the mode-seeking behavior of score-distillation sampling, and strong classifier-free guidance~\cite{ho2022classifier} limit the diversity of the textures they are able to generate~\cite{dreamfusion}. Particularly relevant is TextMesh~\cite{textmesh} which tiles views of a 3D object in a 2x2 grid to enhance consistency in the generated texture. 

Closer to our texturing application are recent approaches~\cite{texfusion, text2tex, TEXTure, HCTM}, that leverage a depth-conditioned Stable Diffusion~\cite{StableDiffusion, ControlNet} to directly project generated pixels on the mesh using a few passes through the diffusion model~\cite{text2room}. 
The main challenges for these approaches is to ensure consistency between the pixels being projected on the mesh from different views. 
TEXTure~\cite{TEXTure} introduces an iterative approach, where each camera view point is used iteratively to refine and generate the texture in the missing regions. TexFusion~\cite{texfusion} 
interleaves diffusion steps and projection steps to improve consistency between the different views.  
Text2Tex~\cite{text2tex} introduces a next best view selection scheme to select the best camera viewpoint given the current coverage of the texture. HCTM~\cite{HCTM} improves consistency by employing textual inversion on the first generated image as well as performing a LoRA-finetuning of the backbone diffusion model, though this comes at the cost of added runtime \ie overall 35 minutes. 
Wang \etal~\cite{breathingnewlife} use a pipeline leveraging some of these ideas to texture all of the 12K objects from over 270 categories of ShapeNetSem~\cite{shapenet}, which shows the generality and potential of projecting generated 2d pixels in 3D. 
Compared to those approaches, we perform better consistency through a grid pattern diffusion, and avoid texture bleeding between different parts by conditioning on both depth and edges. Furthermore, we go beyond RGB texturing via material retrieval. This ensures high texture resolution, removes artifacts due to baked in lighting, and produces relightable assets.

\vspace{3mm}
\noindent\textbf{Concurrent work.}
At the time of this submission, different approaches have been proposed on Arxiv for 3D geometry and texture generation. DreamCraft3D~\cite{sun2023dreamcraft3d} proposes to jointly fine-tune a 2D diffusion model using Dreambooth~\cite{ruiz2022dreambooth} while using it in an SDS optimization to refine a mesh texture. TextureDreamer~\cite{yeh2024texturedreamer} also uses a finetuned diffusion model based on a few reference images and performs SDS type of optimization to optimize BRDF parameters that represent the appearance of an object. In these methods, both the finetuning and the SDS optimization parts that are repeated for each input mesh take significant amount of time. Paint-it~\cite{youwang2023paintit} presents a similar SDS optimization strategy to optimize for the physically based rendering maps that are represented via a neural networks. Despite high-quality results, this approach also requires a large number of forward passes through a diffusion model and can be prohibitively expensive (e.g., 15-30 minutes per mesh), further the available lighting signal available for optimizing specular components of the materials are sparse and potentially inconsistent, making the optimization for materials challenging. Finally, FlashTex~\cite{deng2024flashtex} extends such an optimization process with a light conditioned ControlNet. In contrast, our method avoids such a costly optimization process and utilizes high quality procedural materials. In addition to relighting, this also enables to further edit the appearance of the objects which sets our method apart from concurrent work.

%% file: sec/3_method.tex
\section{Method}
\label{sec:method}

\begin{figure}[t!]
    \centering
    \begin{overpic}[width=\columnwidth]{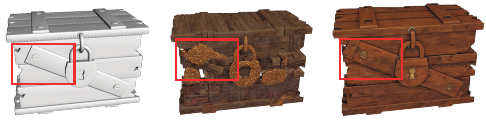}
        \put(9,-3){
            \color{black} Input mesh
        }
        \put(39,-3){
            \color{black} Cond w/ depth
        }
        \put(69,-3){
            \color{black} Cond w/ depth \& lines
        }
    \end{overpic}
    \vspace{2mm}
    \caption{
    \textbf{Conditional generation.}
    We utilize both depth and lineart based conditioning to guide the image generation model. While depth helps to preserve the underlying geometry, occluded and suggested contours represented in the line renderings help to capture details and avoid texture bleeding across different parts.}
    \label{fig:cond}
\end{figure}

The goal of our method is to leverage a pre-trained large scale 2D image generation model to texture a 3D mesh based on a target text prompt. Specifically, we use a diffusion model that is guided with additional control signals that we derive from the underlying 3D model (Section~\ref{sec:preliminary}). In order to ensure synthesis of consistent images across multiple views, we use a cross-frame attention strategy (Section~\ref{sec:constview}). Specifically, we tile multiple views in a grid-like fashion and run a single image generation call that synthesizes the multiple views in a synchronized manner. Finally, we run our method in multiple passes both to refine the texture quality and to ensure full coverage (Section~\ref{sec:refinement}). Given a consistently textured 3D model and access to a material database, we also propose a method to automatically retrieve and assign appropriate materials to different parts of the mesh to output a fully relightable asset (Section~\ref{sec:material}). Next, we discuss each stage of our method in more details.

\subsection{Conditioned image generation}
\label{sec:preliminary}

Our method uses a pretrained large scale image diffusion model to construct a texture for a 3D model based on a text prompt. In our experiments, we leverage Stable Diffusion~\cite{StableDiffusion}, however, our method can be applied to other image diffusion models. Given the success of such models in generating very high quality images, many different additional control signals have been proposed to guide the generation process. We leverage such control mechanisms, namely ControlNet~\cite{ControlNet}, to generate images that are faithful to the underlying geometry of the 3D model. Specifically, we use both depth and lineart conditioned generation. Depth conditioning is crucial to generate images that are aligned with the underlying geometry. However, since the depth is normalized to a predefined range before being passed through the diffusion model, a loss in geometry resolution is inevitable, erasing small details. To address this challenge and to further avoid color bleeding across different pronounced parts of the 3D model, we use the additional lineart conditioning. Specifically, we first render the 3D model with a constant diffuse color~\cite{phong1998illumination} and detect occluded and suggestive contours using the ridge detection filterr~\cite{Lee2007LineDV, decarlo, pearson} from OpenCV~\cite{opencv_library} as additional guidance. Figure~\ref{fig:cond} visualizes how each control signal contributes to the quality of the generated images.

\subsection{Cross-frame attention for consistent image generation}
\label{sec:constview}
A typical image diffusion model consists of cross and self attention blocks. While cross attention models the relation between conditioning signals such as text prompts, self attention layers influence the overall structure and appearance of the generated images. Recent work (some of which are not published yet during the time of submission) adapts image generation models for video editing~\cite{pix2video,geyer2023tokenflow} or multi-view synthesis~\cite{shi2023MVDream} have shown that adapting the self attention layers to perform \emph{cross-frame} attention is critical for consistent appearance synthesis. A straightforward approach to introduce cross frame attention is to stack multiple views in a single image in a grid-like fashion (16 images arranged in a 4-by-4 grid in our experiments) and perform a single image generation process (see Figure~\ref{fig:texture}). At each spatial location in the grid image, the query features, i.e., the features of the corresponding view, attend to the key and value features of the whole grid image, i.e., a union of the self attention features of each view.

We empirically observe that generated images with white background tend to have more uniform lighting compared to images with cluttered background. To enforce white background in the generated views, we follow an approach similar to Blended Diffusion~\cite{blended_diffusion}. We encode a pure white image that has the same size as our grid image into the latent space of Stable Diffusion. We add noise to the resulting latent representation (based on the first diffusion timestep $t=980$ in our experiments) to obtain $z_w^t$. We blend $z_w^t$ with pure Gaussian noise, $z^t$, based on a mask image $M$ constructed by arranging the rendered masks of the 3D model from multiple views in a similar grid. In practice, we dilate $M$ (where white denotes the object region and black denotes the background) before blending to avoid unnecessary white color leaking into the texture. We initialize the diffusion process with such blended noise: $\Tilde{z}^t = (1-M) \odot z_w^t + M \odot z^t$.

\subsection{Multi-pass texture refinement}
\label{sec:refinement}

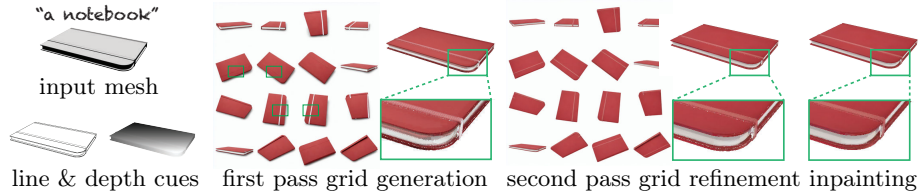
\begin{figure}[ht!]
    \centering
    \begin{overpic}[width=\textwidth]{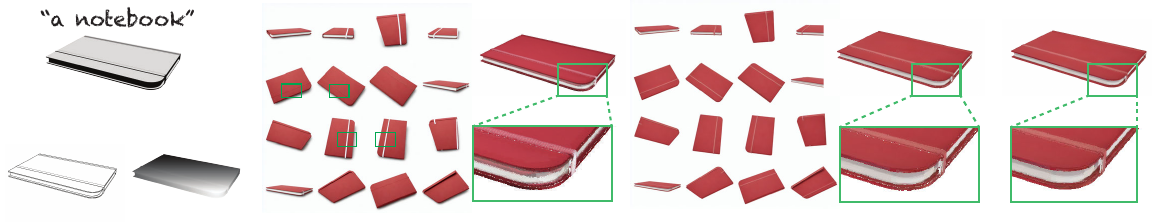}
        \put(3,9){
            \color{black} input mesh
        }
        \put(0,-1){
            \color{black} line \& depth cues
        }
        \put(23,-1){
            \color{black} first pass grid generation
        }
        \put(54,-1){
            \color{black} second pass grid refinement
        }
        \put(87,-1){
            \color{black} inpainting
        }
    \end{overpic}
    \caption{
    \textbf{Multi-pass texture generation.}
    We perform a multi-pass texture generation where an initial pass enables the generation of a globally consistent but potentially blurry texture. In the second pass, we refine this texture quality. We perform a final inpainting pass to ensure a full coverage of any untextured parts of the model. We use lineart and depth cues to condition the image generation model.}
    \label{fig:texture}
    \vspace{-1.2em}
\end{figure}

The cross frame attention strategy described in the previous section ensures global consistency in the synthesized views. However, we observe that small inconsistencies might remain due to limited resolution. Furthermore, different light effects in the generated images across viewpoints can result in color changes. To address such issues, we propose a multi-pass texture refinement process as described below and shown in Figure~\ref{fig:texture}. 

\textbf{First global consistency pass}. In the first pass, we generate a grid image of multiple views starting from full noise as described in Section~\ref{sec:constview}. For this step, we normalize the input 3D model into a unit sphere and sample views at three different elevation levels with uniformly sampled azimuth angles. We then blend the generated views into the texture space using averaging. Averaging helps to blend the small inconsistencies due to the baked lighting across
multiple views so the texture has a more flat lighting overall. This also results in a more blurry texture, however, which we refine in the subsequent stage. 

\textbf{Texture refinement pass}. In the second pass, we render the model with the initial texture obtained from the first pass from the same viewpoints arranged in a grid and re-generate an image. Due to the average blending, the texture and hence the rendered grid images are potentially blurry. To refine these images, we add partial noise and denoise similar to SDEdit~\cite{meng2022sdedit} (in our experiments we perform $20$ steps of denoising where a full denoising is defined to be $50$ steps). Performing a small number of denoising steps helps to generate sharper images without sacrificing the consistency. We then update the texture by blending the refined views. At this stage instead of average blending, for each triangle of the input mesh, we grab its texture from the view where the viewing direction best aligns with the normal of the triangle. 

\textbf{Texture completion pass}. Since the initial viewpoints we sample around the object may not guarantee full coverage, we perform a final pass where we sample additional viewpoints to reveal the parts of the object which may not have been textured yet. 
We therefore iterate over 150 possible viewpoints, sampled on the Fibonacci sphere, and sort them by the amount of new UV space information they provide. We define new information as either a significantly better view for a previously textured pixel or observation of a previously untextured pixel. We define a \emph{contribution} score for each viewpoint based on the new information it brings. 

We then select the most informative viewpoint and recompute contribution of the remaining viewpoints. We repeat this process until the contribution score of the most informative viewpoint falls below a threshold. We refer to the supplementary material for details of how the contribution score and the threshold is defined. In our experiments, we observe that typically $2-3$ additional viewpoints are selected in this pass. 
Given the additional views, we generate the corresponding images iteratively while using blended diffusion to perform inpainting --ensuring consistency with pre-existing texture-- with depth and edge conditioning~\cite{TEXTure}. 
In cases where parts of the mesh occludes others (e.g., a book occluding parts of the shelf), it may still be challenging to obtain full coverage. Hence, we perform a final texture inpainting pass using PatchMatch~\cite{Barnes:2009:PAR}.

\begin{figure*}[t!]
    \centering
    \begin{overpic}[width=\textwidth]{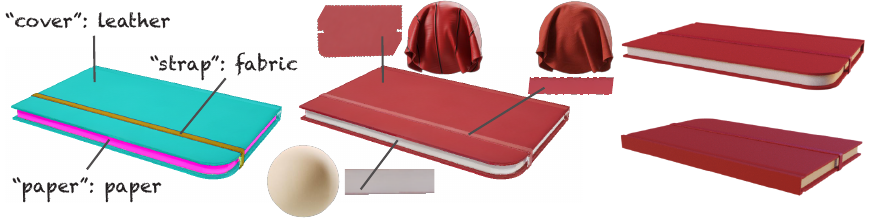}
        \put(-1,-1){
            \color{black} (a) per-part material types
        }
        \put(38,-1){
            \color{black} (b) retrieved materials
        }
        \put(68,-1){
            \color{black} (c) rendered with materials
        }
    \end{overpic}
    
    \caption{
    \textbf{Material retrieval.}
    (a) Given a textured and segmented object, we rely on large language models to extract the global context and suggest different material types for each part. (b) We then perform a visual material search for each part within the corresponding material category. This search utilizes CLIP image embeddings as well as color features. (c) After assinging the retrieved materials to the mesh, we can relight the asset. Here we render it with Blender~\cite{blender} under a soft environment illumination.}
    \label{fig:mat}
    \vspace{-1.2em}
\end{figure*}

\subsection{Material retrieval and assignment}
\label{sec:material}
In many applications (\eg architecture, video-games, product staging), having a fully relightable asset is needed. To address this need, we propose a novel material retrieval and assignment process, leveraging existing databases of high quality materials~\cite{AmbientCG, Source:2023, PolyHaven, Deschaintre20}.

\textbf{Material-aware segmentation}. For any such retrieval process, the first task is to segment the input 3D object into parts where each part can be assigned the same material. For many 3D repositories (e.g., Objaverse~\cite{objaverse}), objects already come with such information (e.g., having part-level semantic segmentation or having disconnected parts). In cases where this information is not given, we use an interactive semi-automatic segmentation method. In particular, we let the user draw a few scribbles on the mesh to indicate material regions to segment.

\textbf{Material search}. Given the generated texture and segments of the 3D model, our core material retrieval algorithm leverages both global context as well as local visual cues. Relying only on visual cues can be ambiguous in determining material types since two very different material types might have a similar appearance. As humans we rely on our prior knowledge of what material types are often suitable for different objects and their parts (e.g., a cushion of an armchair is likely to be made of fabric). Hence, we propose an approach where we first leverage large language models to extract global context information. Specifically, given an object and its different parts, we ask GPT-4V~\cite{chatgpt} to suggest different material categories corresponding to different parts of the object as shown in the inset figure. Then, we render each part of the object from a viewpoint that is as frontal as possible and crop a square patch from this rendering. For each part, we perform a material search based on the visual cues provided by such a patch among the materials in the database that belong to the same category as the suggestion from GPT-4V. 

\begin{wrapfigure}[10]{r}{0.35\columnwidth}
  \begin{center}
    \includegraphics[width=0.35\columnwidth]{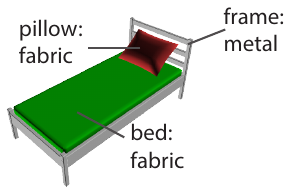}
  \end{center}
\end{wrapfigure}

To perform visual material search we combine CLIP feature similarity with color histograms. CLIP captures texture and semantic concepts and has been shown to work better than other metrics for material retrieval from images~\cite{deschaintre23_visual_fabric, Yan:2023:PSDR-Room}. However, it is not as sensitive to color, sometimes resulting in color differences between the RGB texture and the retrieved material. Hence, we augment it with LAB space color information.
In a one-time pre-processing stage, we render each material in our database on a flat surface with a neutral environment illumination. We compute CLIP image embedding features as well as color histogram for each material rendering. The color histograms are 3D histograms in the La*b* space with 8 bins for L and 32 for a* and b*. In particular we empirically found that extracting the 7 most prominent colors is a compact and sufficient representation for high quality retrieval. Given the patch of the rendering of the part, we compute the corresponding CLIP image embedding and color histogram in a similar manner. We compute the distance of the patch to all the materials of the specific category both based on the CLIP similarity ($d_{clip}$) as well as the color histograms ($d_{color}$). We normalize both $d_{clip}$ and $d_{color}$ to be in the range $(0,1)$ based on the maximum distance between the patch and all other materials. The final distance is a linear combination of the two distances: $d = d_{clip} * (1-w) + d_{color} * w$ where $w$ (set to $0.8$ in our experiments) is a weight that controls how much the search should be influenced by color similarity. We assign the material that has the smallest distance $d$ to each part.

\textbf{Material assignment.} 
When assigning the selected materials to the mesh parts, we adjust the material scale based on the physical size of the materials and the mesh parts. The information about the mesh physical size is often available in high quality asset libraries, but can also be approximated leveraging GPT-4V~\cite{chatgpt}. We compute the mesh and uv area of each part, and apply a tiling factor to the selected material so that it matches the physical size of the part. 
The height maps are also normalized across materials, in order to get a consistent mid-level and meaningful relative displacement values. 
To minimize the amount of seams when texturing the mesh with the materials, we use a triplanar projection onto the mesh instead of a simple uv projection.
Finally, the materials can be blended together to obtain a single texture set for the entire object, or kept separately to allow for further editing of each part. 

%% file: sec/4_evaluation.tex
\section{Evaluation}
\label{sec:evaluation}

\subsection{Implementation Details}
For texture generation, we use Stable Diffusion v1.5~\cite{SD-web} as the base model with depth and lineart ControlNet models~\cite{ControlNet} for conditioning. We set the weight of each control channel to $0.5$ in our experiments. In the first two passes we generate images in $1600$-by-$1600$ resolution where generation with full denoising ($50$ steps) in the first pass takes around $54$ seconds and partial denoising ($20$ steps) in the second pass takes $21$ seconds. In the final pass, we generate each additional viewpoint in $512$-by-$512$ resolution where each generation takes $4.4$ seconds and PatchMatch based inpainting takes 1 second. Our output textures are of resolution $1024$-by-$1024$. In total, our texturing step takes around $2.5-3$ minutes on a A100 GPU. Selecting the best viewpoint to render a representative patch for each part and performing part-based material retrieval typically takes several seconds.

For the material retrieval results shown in this paper, we use the Substance 3D Assets material library~\cite{Source:2023} containing $8,965$ procedural materials across $16$ categories (e.g., wood, metal, fabric, paper etc.). For each material we extract the different presets (typically 1 to 4) defined by the artist in the procedural material itself resulting in $28,206$ variations in total. 

\begin{table}
\begin{center}
 \caption{
 \textbf{Quantitative comparison.}}
\footnotesize
\begin{tabular}{cccccc}
\toprule
 &Text2Tex~\cite{text2tex} & TEXTure~\cite{TEXTure}  & Ours w/o line guidance & Ours w/ inpainting & Ours \\
\midrule
FID $\downarrow$ & 42.133 & 49.276 & 46.958 & 38.535 & \textbf{38.467} \\
 \bottomrule
 \end{tabular}
 \label{tab:quanti_result}
 \end{center}
\end{table}

\subsection{Evaluation of texture generation.}
\begin{figure}[h!]
    \begin{center}
        \includegraphics[width=\textwidth]{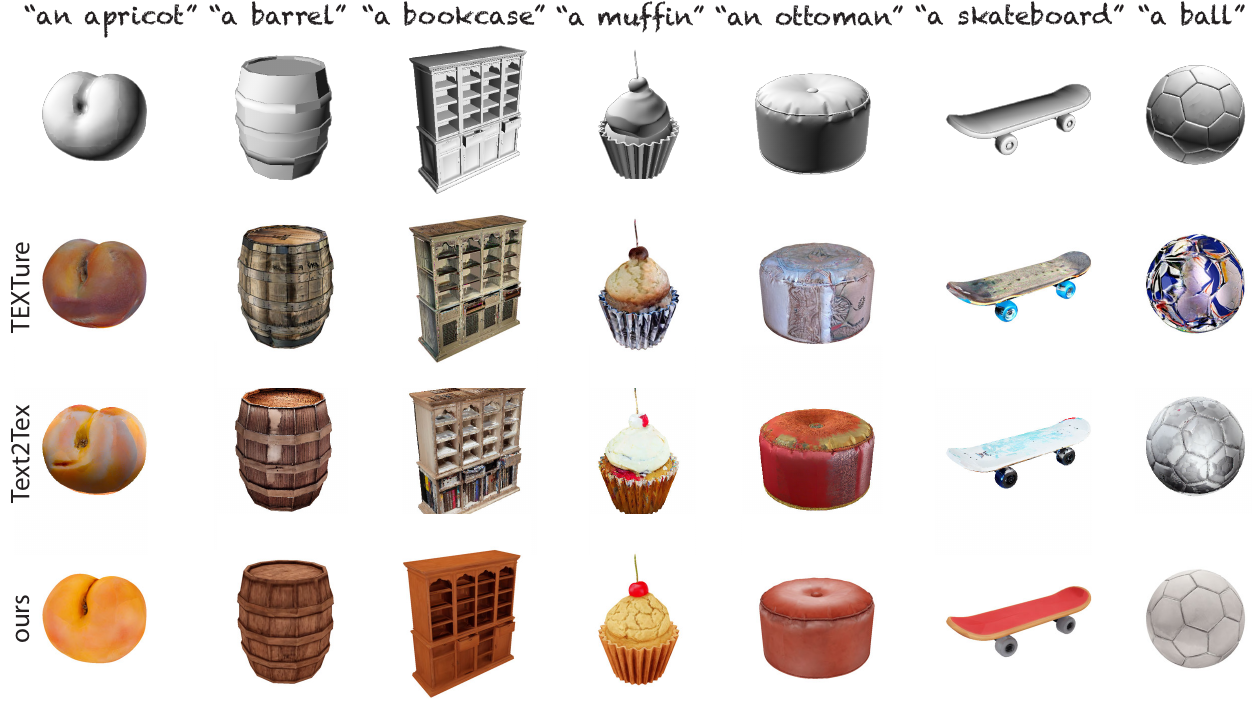}
        \caption{We compare our method to Text2Tex~\cite{text2tex} and TEXTure~\cite{TEXTure}, two recent state-of-the-art generative texturing methods that also utilize Stable Diffusion. We show the input meshes with the text prompts in the first row. Our results better preserve texture consistency.}
        \label{fig:tex_results}
    \end{center}%
\end{figure}

In order to evaluate the performance of the texture generation part of our method, we use the subset of the Objaverse dataset~\cite{objaverse} that contains $410$ models as proposed by Text2tex~\cite{text2tex} along with the captions that are derived from the names of the models. We compare our method to the recent generative texturing methods that also utilize Stable Diffusion as the backbone and have provided code and data. Specifically, we compare to TEXTure~\cite{TEXTure} and Text2Tex~\cite{text2tex} which both utilize depth conditioned generation with strategies to minimize seams. We note that both methods report superior performance against other CLIP-based optimization methods~\cite{clipmesh,text2mesh}. As a quantitative metric, we adopt the commonly used Frechet Inception Distance (FID)~\cite{fid_metric} to evaluate the quality of the textures. Following the protocol introduced by Text2Tex, we use the renderings of the Objaverse models with ground truth textures for the real distribution. 
We provide the quantitative results in Table~\ref{tab:quanti_result} and provide visual results in Figure~\ref{fig:tex_results}. As shown both visually and quantitatively, our method demonstrates superior performance with respect to the related work. 

\textbf{User Study.} We conduct a user study where we select 5 random pairs of results from our method and each of the baseline methods. We show a 360-video of each result rendered with similar lighting. We ask the users which result looks better overall considering realism and texture seams. We collect $140$ responses from $28$ users for each method comparison. Our method is preferred $67.86$\% of the time ($95$\% confidence interval: $\pm8.9$\%) with respect to TEXTure and $60$\%  of the time ($95$\% confidence interval: $\pm10.1$\%) with respect to Text2Tex. We refer to the supplementary material for the details of the user study.

\begin{figure}[h!]
    \begin{center}
        \includegraphics[width=\linewidth]{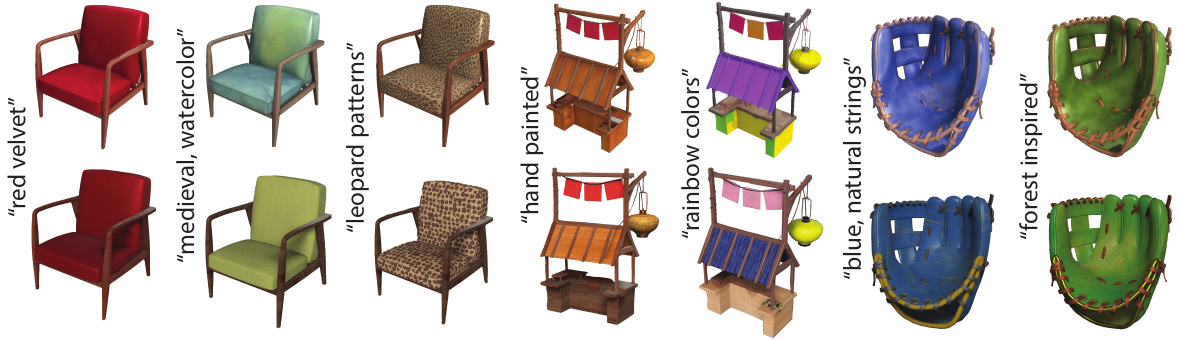}
        \caption{For the same model, we show the texturing (top row) and material retrieval (bottom row) variations obtained by using different input prompts.}
        \label{fig:variations}
    \end{center}%
\end{figure}

Finally, we also provide both texturing and material retreival variations in Figure~\ref{fig:variations} to show how our method generalizes to different text prompts.

\textbf{Ablation Study.} We ablate different design decisions in our method. Specifically, we run our texturing method using only depth conditioned generation (Table~\ref{tab:quanti_result}, column 3), using depth and line conditioned generation without the final patch match based inpainting (Table~\ref{tab:quanti_result}, column 4), and the full pipeline. As shown in the table, using line art conditioning significantly improves the performance since it provides a strong structural guidance. While the final inpainting does not have a significant impact on the quantitative metrics, we visually observe that it improves the quality as shown in the supplementary material.

In Figure~\ref{fig:ablation:blendingtexture}, we ablate the effect of using multiple texturing passes on the resulting texture quality. Specifically, we show the texturing results obtained by (i) using a single pass with view-based blending, and finally (ii) our proposed multi-pass approach. As seen in the figure, our proposed approach reduces seams and generates higher quality textures.

\label{sec:ablation}
\begin{figure}[ht]
    \begin{center}
    \begin{overpic}[width=\linewidth]{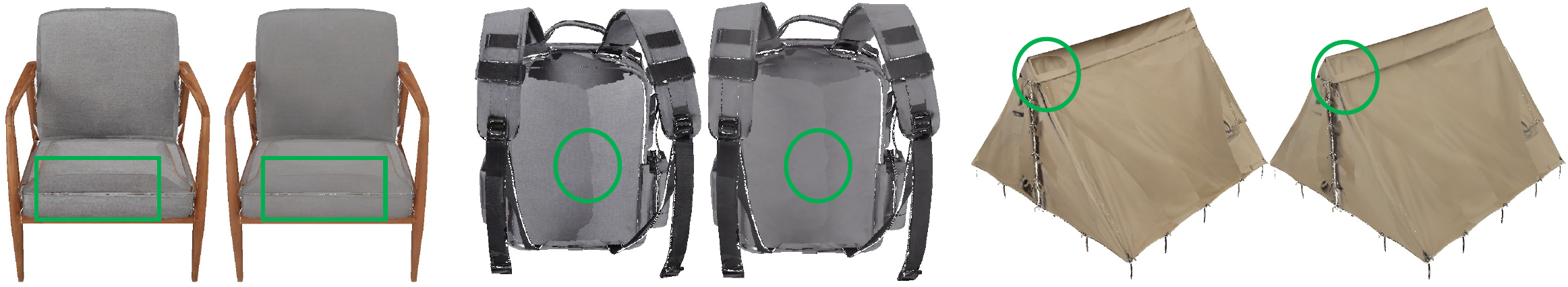}
        \put(3,-1){
            \color{black} single
        }
        \put(17,-1){
            \color{black} multi
        }
        \put(33,-1){
            \color{black} single
        }
        \put(48,-1){
            \color{black} multi
        }
        \put(67,-1){
            \color{black} single
        }
        \put(87,-1){
            \color{black} multi
        }
    \end{overpic}
        \caption{Compared to a single pass with view-based texture blending, our multi-pass approach reduces texture seams.}
        \label{fig:ablation:blendingtexture}
    \end{center}%
\end{figure}

\subsection{Evaluation of material assignment.} 
\begin{figure}[t!]
    \begin{center}
        \includegraphics[width=\textwidth]{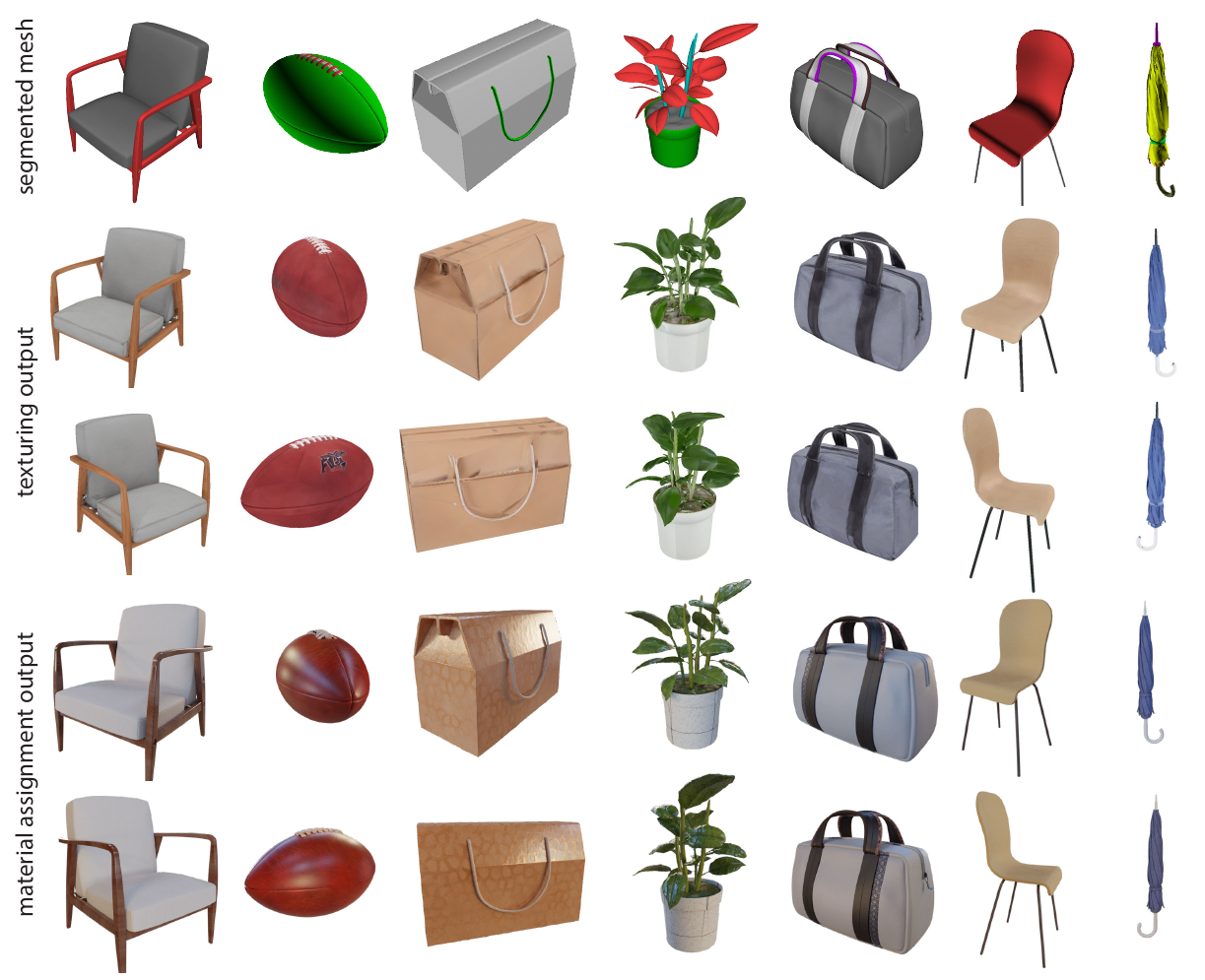}
        \caption{Given an input segmented mesh from the Source 3D assets (top row, each segment is shown in a different color), we show results from our automatic texturing (second and third rows) and the renderings obtained by matching the textures to closest materials in our database (fourth and fifth rows).}
        \label{fig:mat_results}
    \end{center}%
\end{figure}
We evaluate our material assignment pipeline on 3D models obtained from the Adobe Substance 3D Assets library~\footnote{https://substance3d.adobe.com/assets/} which contains models across various categories (e.g., tools, technology, building etc.) The 3D models in this database are segmented with available semantic part names which we utilize to query ChatGPT. In Figure~\ref{fig:mat_results}, we show representative qualitative results. We test our material retrieval pipeline on a subset of Objaverse models where we use the part segmentations already provided by the model and material query suggestions obtained from ChatGPT-4V. Finally, we also demonstrate that our texturing and material assignment pipeline can be applied to objects from the Google Scanned Objects dataset~\cite{Downs2022}. In this case, we use our proposed semi-automatic texture guided segmentation approach. We show results in Figure~\ref{fig:wild_segmentation} and refer to the supplementary material for more details and results. In the supplementary, we also show relighting and editing results.

\begin{figure}[h!]
    \begin{center}
        \includegraphics[width=\linewidth]{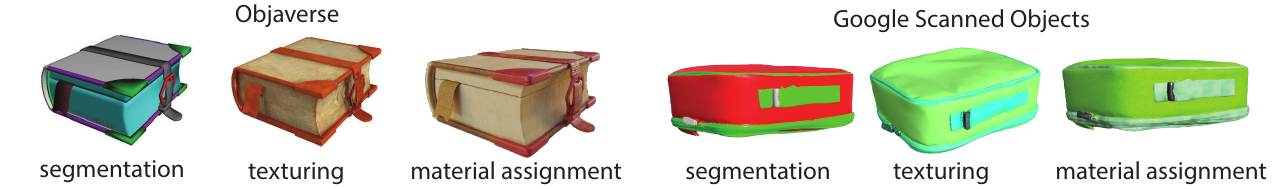}
        \caption{We evaluate our method on 3D models from the Objaverse dataset which come as disconnected components as well as scanned models in the Google Scanned Objects dataset where we use a semi-automatic segmentation method.}
        \label{fig:wild_segmentation}
    \end{center}%
\end{figure}

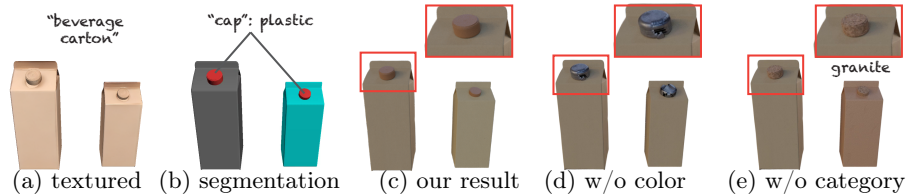
\begin{figure*}[t!]
    \centering
    \begin{overpic}[width=\textwidth]{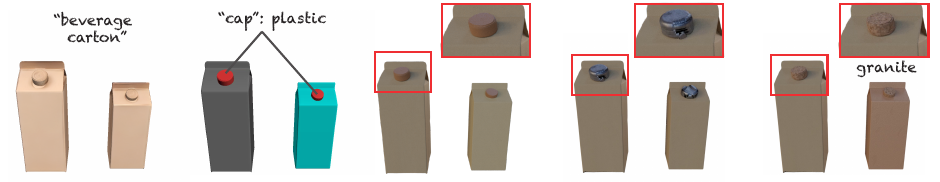}
        \put(1,-1){
            \color{black} (a) textured
        }
        \put(17,-1){
            \color{black} (b) segmentation
        }
        \put(41,-1){
            \color{black} (c) our result
        }
        \put(59,-1){
            \color{black} (d) w/o color
        }
        \put(79,-1){
            \color{black} (e) w/o category
        }
    \end{overpic}
    
    \caption{
    \textbf{Material retrieval comparison.}
    (a) Given a textured mesh, (b) we rely on LLMs to get material category suggestions for each segment of the object. (c) We then perform category specific material retrieval using visual cues. (d) If we do not use explicit color similarity for search, we observe that retrieval results are not satisfactory especially for small regions. (e) While omitting material category suggestions might result in visually similar material retrieval results, they often do not correspond to realistic suggestions.}
    \label{fig:mat_abl}
    \vspace{-1.2em}
\end{figure*}

We ablate certain design choices in our material search algorithm. First, instead of using a combination of CLIP image and color similarity, we use only CLIP similarity as our visual metric (w/o color). Next, instead of category specific search, we perform only visual search within the whole material database (w/o category). As shown in Figure~\ref{fig:mat_abl}, we find our method provides a good balance between plausibility and quality. When working with locally cropped patches from different parts of the 3D model, the CLIP image similarity may not be as sensitive to color, sometimes resulting in color differences between the RGB texture and the retrieved material. Omitting the category label in the search results in unrealistic material suggestions. We provide further discussion in the supplementary.

%% file: sec/5_conc.tex
\section{Conclusion}
\label{sec:conc}
We present a method to generate relightable textures for 3D models given a text prompt. Our method uses a pre-trained text-to-image generation model in the first stage to generate a consistent and high quality RGB texture. We then use this texture in conjunction with priors from a large language model to retrieve materials from a material database to assign to different parts of the model. We compare our texturing results against state of the art and demonstrate superior performance. We also show that unlike previous work, our method provides fully relightable and editable outputs, highly desired in many applications.

While promising, our method also has limitations of interest for future work. While the cross-frame attention strategy greatly improves the consistency of the textures, we still observe that generations across different views might differ in the appearance of the details. Recent methods have demonstrated successful results in finetuning the image generation models to output consistent multi view images~\cite{zero123, shi2023zero123}. Using such finetuned models along with depth and line art guidance is a promising direction to pursue. 
While we can output high quality and relightable assets by matching the generated textures to materials in a database, we are not able to capture certain details like logos or dirt that exist in the generated textures. Recovering such details as an additional layer is an interesting future direction.

%% file: sec/6-ackowledgements.tex
\section{Acknowledgments}
\label{sec:Acknowledgments}

We thank Aaron Hertzmann for sharing his vast expertise on suggestive contours, and sharing his code snipped for the ridge detection filter.

%% file: main.bbl
\begin{thebibliography}{10}
\providecommand{\url}[1]{\texttt{#1}}
\providecommand{\urlprefix}{URL }
\providecommand{\doi}[1]{https://doi.org/#1}

\bibitem{SD-web}
Stable diffusion. \url{https://github.com/ Stability-AI/stablediffusion}, accessed: 2023

\bibitem{Source:2023}
Adobe: {Substance 3D Assets}. \url{https://substance3d.adobe.com/assets/} (2023)

\bibitem{AmbientCG}
{AmbientCG}: \url{https://www.ambientcg.com/} (2023)

\bibitem{blended_diffusion}
Avrahami, O., Lischinski, D., Fried, O.: Blended diffusion for text-driven editing of natural images. In: Proceedings of the IEEE/CVF Conference on Computer Vision and Pattern Recognition (CVPR). pp. 18208--18218 (June 2022)

\bibitem{Barnes:2009:PAR}
Barnes, C., Shechtman, E., Finkelstein, A., Goldman, D.B.: {PatchMatch}: A randomized correspondence algorithm for structural image editing. ACM Transactions on Graphics (Proc. SIGGRAPH)  \textbf{28}(3) (Aug 2009)

\bibitem{Berkiten17}
Berkiten, S., Halber, M., Solomon, J., Ma, C., Li, H., Rusinkiewicz, S.: Learning detail transfer based on geometric features. Computer Graphics Forum  \textbf{36},  361--373 (05 2017). \doi{10.1111/cgf.13132}

\bibitem{opencv_library}
Bradski, G.: {The OpenCV Library}. Dr. Dobb's Journal of Software Tools  (2000)

\bibitem{texfusion}
Cao, T., Kreis, K., Fidler, S., Sharp, N., Yin, K.: Texfusion: Synthesizing 3d textures with text-guided image diffusion models. In: Proceedings of the IEEE/CVF International Conference on Computer Vision. pp. 4169--4181 (2023)

\bibitem{pix2video}
Ceylan, D., Huang, C.H., Mitra, N.J.: Pix2video: Video editing using image diffusion (2023)

\bibitem{shapenet}
Chang, A.X., Funkhouser, T., Guibas, L., Hanrahan, P., Huang, Q., Li, Z., Savarese, S., Savva, M., Song, S., Su, H., Xiao, J., Yi, L., Yu, F.: {ShapeNet: An Information-Rich 3D Model Repository}. Tech. Rep. arXiv:1512.03012 [cs.GR], Stanford University --- Princeton University --- Toyota Technological Institute at Chicago (2015)

\bibitem{text2tex}
Chen, D.Z., Siddiqui, Y., Lee, H.Y., Tulyakov, S., Nie{\ss}ner, M.: Text2tex: Text-driven texture synthesis via diffusion models. In: Proceedings of the IEEE/CVF International Conference on Computer Vision (ICCV). pp. 18558--18568 (October 2023)

\bibitem{fantasia3d}
Chen, R., Chen, Y., Jiao, N., Jia, K.: Fantasia3d: Disentangling geometry and appearance for high-quality text-to-3d content creation. arXiv preprint arXiv:2303.13873  (2023)

\bibitem{Chen:2012:NTT}
Chen, X., Funkhouser, T., Goldman, D.B., Shechtman, E.: Non-parametric texture transfer using {MeshMatch}. Adobe Technical Report 2012-2  (Nov 2012)

\bibitem{blender}
Community, B.O.: Blender - a 3D modelling and rendering package. Blender Foundation, Stichting Blender Foundation, Amsterdam (2018), \url{http://www.blender.org}

\bibitem{decarlo}
DeCarlo, D., Finkelstein, A., Rusinkiewicz, S., Santella, A.: Suggestive Contours for Conveying Shape. Association for Computing Machinery, New York, NY, USA, 1 edn. (2023), \url{https://doi.org/10.1145/3596711.3596755}

\bibitem{objaverse}
Deitke, M., Schwenk, D., Salvador, J., Weihs, L., Michel, O., VanderBilt, E., Schmidt, L., Ehsani, K., Kembhavi, A., Farhadi, A.: Objaverse: A universe of annotated 3d objects. In: Proceedings of the IEEE/CVF Conference on Computer Vision and Pattern Recognition. pp. 13142--13153 (2023)

\bibitem{deng2024flashtex}
Deng, K., Omernick, T., Weiss, A., Ramanan, D., Zhu, J.Y., Zhou, T., Agrawala, M.: Flashtex: Fast relightable mesh texturing with lightcontrolnet. arXiv preprint arXiv:2402.13251  (2024)

\bibitem{Deschaintre20}
Deschaintre, V., Drettakis, G., Bousseau, A.: Guided fine-tuning for large-scale material transfer. Computer Graphics Forum (Proceedings of the Eurographics Symposium on Rendering)  \textbf{39}(4) (2020), \url{http://www-sop.inria.fr/reves/Basilic/2020/DDB20}

\bibitem{deschaintre23_visual_fabric}
Deschaintre, V., Guerrero-Viu, J., Gutierrez, D., Boubekeur, T., Masia, B.: The visual language of fabrics. ACM Trans. Graph.  (2023)

\bibitem{dong2007optimized}
Dong, W., Zhou, N., Paul, J.C.: Optimized tile-based texture synthesis. In: Proceedings of Graphics Interface 2007. pp. 249--256 (2007)

\bibitem{Downs2022}
Downs, L., Francis, A., Koenig, N., Kinman, B., Hickman, R., Reymann, K., McHugh, T.B., Vanhoucke, V.: Google scanned objects: A high-quality dataset of 3d scanned household items. In: 2022 International Conference on Robotics and Automation (ICRA). p. 2553–2560. IEEE Press (2022). \doi{10.1109/ICRA46639.2022.9811809}, \url{https://doi.org/10.1109/ICRA46639.2022.9811809}

\bibitem{efros2001image}
Efros, A.A., Freeman, W.T.: Image quilting for texture synthesis and transfer. In: Proceedings of the 28th annual conference on Computer Graphics and Interactive Techniques. pp. 341--346 (2001)

\bibitem{efros1999texture}
Efros, A.A., Leung, T.K.: Texture synthesis by non-parametric sampling. In: Proceedings of the IEEE/CVF International Conference on Computer Vision (ICCV). vol.~2, pp. 1033--1038. IEEE (1999)

\bibitem{textdeformer}
Gao, W., Aigerman, N., Thibault, G., Kim, V., Hanocka, R.: Textdeformer: Geometry manipulation using text guidance. In: ACM Transactions on Graphics (SIGGRAPH) (2023)

\bibitem{geyer2023tokenflow}
Geyer, M., Bar-Tal, O., Bagon, S., Dekel, T.: Tokenflow: Consistent diffusion features for consistent video editing. In: arXiv (2023)

\bibitem{537718}
Heeger, D., Bergen, J.: Pyramid-based texture analysis/synthesis. In: Proceedings., International Conference on Image Processing. vol.~3, pp. 648--651 vol.3 (1995). \doi{10.1109/ICIP.1995.537718}

\bibitem{fid_metric}
Heusel, M., Ramsauer, H., Unterthiner, T., Nessler, B., Hochreiter, S.: Gans trained by a two time-scale update rule converge to a local nash equilibrium. In: Proceedings of the 31st International Conference on Neural Information Processing Systems. p. 6629–6640. NIPS'17, Curran Associates Inc., Red Hook, NY, USA (2017)

\bibitem{ho2022classifier}
Ho, J., Salimans, T.: Classifier-free diffusion guidance. arXiv preprint arXiv:2207.12598  (2022)

\bibitem{text2room}
H{\"o}llein, L., Cao, A., Owens, A., Johnson, J., Nie{\ss}ner, M.: Text2room: Extracting textured 3d meshes from 2d text-to-image models. In: Proceedings of the IEEE/CVF International Conference on Computer Vision (2023)

\bibitem{Kholgade2014}
Kholgade, N., Simon, T., Efros, A., Sheikh, Y.: 3d object manipulation in a single photograph using stock 3d models. ACM Transactions on Computer Graphics  \textbf{33}(4) (2014)

\bibitem{solidtexturesynthesis}
Kopf, J., Fu, C.W., Cohen-Or, D., Deussen, O., Lischinski, D., Wong, T.T.: Solid texture synthesis from 2d exemplars. ACM Transactions on Graphics (Proceedings of SIGGRAPH 2007)  \textbf{26}(3),  2:1--2:9 (2007)

\bibitem{Kraevoy2003}
Kraevoy, V., Sheffer, A., Gotsman, C.: Matchmaker: Constructing constrained texture maps. ACM Trans. Graph.  \textbf{22}(3),  326–333 (jul 2003). \doi{10.1145/882262.882271}, \url{https://doi.org/10.1145/882262.882271}

\bibitem{kwatra2005texture}
Kwatra, V., Essa, I., Bobick, A., Kwatra, N.: Texture optimization for example-based synthesis. In: ACM Transactions on Graphics (ToG). vol.~24, pp. 795--802. ACM (2005). \doi{10.1145/1073204.1073263}

\bibitem{kwatra2003graphcut}
Kwatra, V., Sch{\"o}dl, A., Essa, I., Turk, G., Bobick, A.: Graphcut textures: image and video synthesis using graph cuts. ACM Transactions on Graphics (ToG)  \textbf{22}(3),  277--286 (2003)

\bibitem{Lee2007LineDV}
Lee, Y., Markosian, L., Lee, S., Hughes, J.F.: Line drawings via abstracted shading. ACM SIGGRAPH 2007 papers  (2007), \url{https://api.semanticscholar.org/CorpusID:40738407}

\bibitem{appearancespace}
Lefebvre, S., Hoppe, H.: Appearance-space texture synthesis. ACM Trans. Graph.  \textbf{25},  541--548 (07 2006). \doi{10.1145/1141911.1141921}

\bibitem{sweetdreamer}
Li, W., Chen, R., Chen, X., Tan, P.: Sweetdreamer: Aligning geometric priors in 2d diffusion for consistent text-to-3d. arXiv preprint arXiv:2310.02596  (2023)

\bibitem{magic3d}
Lin, C.H., Gao, J., Tang, L., Takikawa, T., Zeng, X., Huang, X., Kreis, K., Fidler, S., Liu, M.Y., Lin, T.Y.: Magic3d: High-resolution text-to-3d content creation. In: Proceedings of the IEEE/CVF Conference on Computer Vision and Pattern Recognition. pp. 300--309 (2023)

\bibitem{zero123}
Liu, R., Wu, R., Van~Hoorick, B., Tokmakov, P., Zakharov, S., Vondrick, C.: Zero-1-to-3: Zero-shot one image to 3d object. In: Proceedings of the IEEE/CVF International Conference on Computer Vision. pp. 9298--9309 (2023)

\bibitem{long2023wonder3d}
Long, X., Guo, Y.C., Lin, C., Liu, Y., Dou, Z., Liu, L., Ma, Y., Zhang, S.H., Habermann, M., Theobalt, C., et~al.: Wonder3d: Single image to 3d using cross-domain diffusion. arXiv preprint arXiv:2310.15008  (2023)

\bibitem{lu2007context}
Lu, J., Georghiades, A.S., Glaser, A., Wu, H., Wei, L.Y., Guo, B., Dorsey, J., Rushmeier, H.: Context-aware textures. ACM Transactions on Graphics (TOG)  \textbf{26}(1),  3--es (2007)

\bibitem{meng2022sdedit}
Meng, C., He, Y., Song, Y., Song, J., Wu, J., Zhu, J.Y., Ermon, S.: {SDE}dit: Guided image synthesis and editing with stochastic differential equations. In: International Conference on Learning Representations (2022)

\bibitem{geomcorrelation}
Mertens, T., Kautz, J., Chen, J., Bekaert, P., Durand, F.: Texture transfer using geometry correlation. In: Proceedings of the 17th Eurographics Conference on Rendering Techniques. p. 273–284. EGSR '06, Eurographics Association, Goslar, DEU (2006)

\bibitem{latentnerf}
Metzer, G., Richardson, E., Patashnik, O., Giryes, R., Cohen-Or, D.: Latent-nerf for shape-guided generation of 3d shapes and textures. In: Proceedings of the IEEE/CVF Conference on Computer Vision and Pattern Recognition. pp. 12663--12673 (2023)

\bibitem{text2mesh}
Michel, O., Bar-On, R., Liu, R., Benaim, S., Hanocka, R.: Text2mesh: Text-driven neural stylization for meshes. In: Proceedings of the IEEE/CVF Conference on Computer Vision and Pattern Recognition (CVPR). pp. 13492--13502 (June 2022)

\bibitem{clipmesh}
Mohammad~Khalid, N., Xie, T., Belilovsky, E., Popa, T.: Clip-mesh: Generating textured meshes from text using pretrained image-text models. In: SIGGRAPH Asia 2022 conference papers. pp.~1--8 (2022)

\bibitem{oechsle2019texture}
Oechsle, M., Mescheder, L., Niemeyer, M., Strauss, T., Geiger, A.: Texture fields: Learning texture representations in function space. In: Proceedings of the IEEE/CVF International Conference on Computer Vision. pp. 4531--4540 (2019)

\bibitem{chatgpt}
OpenAI: Gpt-4v(ision) system card. Tech. rep., OpenAI, \url{https://cdn.openai.com/papers/GPTV_System_Card.pdf}

\bibitem{photoshape2018}
Park, K., Rematas, K., Farhadi, A., Seitz, S.M.: Photoshape: Photorealistic materials for large-scale shape collections. ACM Trans. Graph.  \textbf{37}(6) (Nov 2018)

\bibitem{pearson}
Pearson, D., Robinson, J.: Visual communication at very low data rates. Proceedings of the IEEE  \textbf{73}(4),  795--812 (1985). \doi{10.1109/PROC.1985.13202}

\bibitem{phong1998illumination}
Phong, B.T.: Illumination for computer generated pictures pp. 95--101 (1998)

\bibitem{PolyHaven}
{PolyHaven}: \url{https://www.polyhaven.com/} (2023)

\bibitem{dreamfusion}
Poole, B., Jain, A., Barron, J.T., Mildenhall, B.: Dreamfusion: Text-to-3d using 2d diffusion. In: The Eleventh International Conference on Learning Representations (2022)

\bibitem{portilla2000parametric}
Portilla, J., Simoncelli, E.P.: A parametric texture model based on joint statistics of complex wavelet coefficients. International Journal of Computer Vision  \textbf{40}(1),  49--70 (2000)

\bibitem{magic123}
Qian, G., Mai, J., Hamdi, A., Ren, J., Siarohin, A., Li, B., Lee, H.Y., Skorokhodov, I., Wonka, P., Tulyakov, S., et~al.: Magic123: One image to high-quality 3d object generation using both 2d and 3d diffusion priors. arXiv preprint arXiv:2306.17843  (2023)

\bibitem{Dalle2}
Ramesh, A., Dhariwal, P., Nichol, A., Chu, C., Chen, M.: Hierarchical text-conditional image generation with clip latents. arXiv preprint arXiv:2204.06125  \textbf{1}(2), ~3 (2022)

\bibitem{TEXTure}
Richardson, E., Metzer, G., Alaluf, Y., Giryes, R., Cohen-Or, D.: Texture: Text-guided texturing of 3d shapes. In: ACM SIGGRAPH 2023 Conference Proceedings. SIGGRAPH '23, Association for Computing Machinery, New York, NY, USA (2023). \doi{10.1145/3588432.3591503}, \url{https://doi.org/10.1145/3588432.3591503}

\bibitem{StableDiffusion}
Rombach, R., Blattmann, A., Lorenz, D., Esser, P., Ommer, B.: High-resolution image synthesis with latent diffusion models. In: Proceedings of the IEEE/CVF conference on computer vision and pattern recognition. pp. 10684--10695 (2022)

\bibitem{ruiz2022dreambooth}
Ruiz, N., Li, Y., Jampani, V., Pritch, Y., Rubinstein, M., Aberman, K.: Dreambooth: Fine tuning text-to-image diffusion models for subject-driven generation  (2022)

\bibitem{shi2023zero123}
Shi, R., Chen, H., Zhang, Z., Liu, M., Xu, C., Wei, X., Chen, L., Zeng, C., Su, H.: Zero123++: a single image to consistent multi-view diffusion base model. arXiv preprint arXiv:2310.15110  (2023)

\bibitem{shi2023MVDream}
Shi, Y., Wang, P., Ye, J., Mai, L., Li, K., Yang, X.: Mvdream: Multi-view diffusion for 3d generation. arXiv:2308.16512  (2023)

\bibitem{siddiqui2022texturify}
Siddiqui, Y., Thies, J., Ma, F., Shan, Q., Nie{\ss}ner, M., Dai, A.: Texturify: Generating textures on 3d shape surfaces. In: European Conference on Computer Vision. pp. 72--88. Springer (2022)

\bibitem{sun2023dreamcraft3d}
Sun, J., Zhang, B., Shao, R., Wang, L., Liu, W., Xie, Z., Liu, Y.: Dreamcraft3d: Hierarchical 3d generation with bootstrapped diffusion prior (2023)

\bibitem{dreamgaussian}
Tang, J., Ren, J., Zhou, H., Liu, Z., Zeng, G.: Dreamgaussian: Generative gaussian splatting for efficient 3d content creation. arXiv preprint arXiv:2309.16653  (2023)

\bibitem{HCTM}
Tang, Z., He, T.: Text-guided high-definition consistency texture model. arXiv preprint arXiv:2305.05901  (2023)

\bibitem{textmesh}
Tsalicoglou, C., Manhardt, F., Tonioni, A., Niemeyer, M., Tombari, F.: Textmesh: Generation of realistic 3d meshes from text prompts. arXiv preprint arXiv:2304.12439  (2023)

\bibitem{texsynthesisonsurfaces}
Turk, G.: Texture synthesis on surfaces. In: Proceedings of the 28th Annual Conference on Computer Graphics and Interactive Techniques. p. 347–354. SIGGRAPH '01, Association for Computing Machinery, New York, NY, USA (2001). \doi{10.1145/383259.383297}, \url{https://doi.org/10.1145/383259.383297}

\bibitem{Tzur_2009_FlexiStickers}
Tzur, Y., Tal, A.: Flexistickers: Photogrammetric texture mapping using casual images. ACM Trans. Graph.  \textbf{28}(3) (jul 2009). \doi{10.1145/1531326.1531351}, \url{https://doi.org/10.1145/1531326.1531351}

\bibitem{sjc}
Wang, H., Du, X., Li, J., Yeh, R.A., Shakhnarovich, G.: Score jacobian chaining: Lifting pretrained 2d diffusion models for 3d generation. In: Proceedings of the IEEE/CVF Conference on Computer Vision and Pattern Recognition. pp. 12619--12629 (2023)

\bibitem{breathingnewlife}
Wang, T., Kanakis, M., Schindler, K., Van~Gool, L., Obukhov, A.: Breathing new life into 3d assets with generative repainting. In: Proceedings of the British Machine Vision Conference (BMVC). BMVA Press (2023)

\bibitem{Wang:2016}
Wang, T.Y., Su, H., Huang, Q., Huang, J., Guibas, L., Mitra, N.J.: Unsupervised texture transfer from images to model collections. ACM Trans. Graph.  \textbf{35}(6),  177:1--177:13 (2016), \url{http://doi.acm.org/10.1145/2980179.2982404}

\bibitem{prolificdreamer}
Wang, Z., Lu, C., Wang, Y., Bao, F., Li, C., Su, H., Zhu, J.: Prolificdreamer: High-fidelity and diverse text-to-3d generation with variational score distillation. arXiv preprint arXiv:2305.16213  (2023)

\bibitem{wei2009state}
Wei, L.Y., Lefebvre, S., Kwatra, V., Turk, G.: State of the art in example-based texture synthesis. Eurographics 2009, State of the Art Report, EG-STAR pp. 93--117 (2009)

\bibitem{arbitrarymanifold}
Wei, L.Y., Levoy, M.: Texture synthesis over arbitrary manifold surfaces. In: Proceedings of the 28th Annual Conference on Computer Graphics and Interactive Techniques. p. 355–360. SIGGRAPH '01, Association for Computing Machinery, New York, NY, USA (2001). \doi{10.1145/383259.383298}, \url{https://doi.org/10.1145/383259.383298}

\bibitem{Yan:2023:PSDR-Room}
Yan, K., Luan, F., Ha\v{s}an, M., Groueix, T., Deschaintre, V., Zhao, S.: Psdr-room: Single photo to scene using differentiable rendering. In: ACM SIGGRAPH Asia 2023 Conference Proceedings (2023)

\bibitem{yeh2024texturedreamer}
Yeh, Y.Y., Huang, J.B., Kim, C., Xiao, L., Nguyen-Phuoc, T., Khan, N., Zhang, C., Chandraker, M., Marshall, C.S., Dong, Z., et~al.: Texturedreamer: Image-guided texture synthesis through geometry-aware diffusion. arXiv preprint arXiv:2401.09416  (2024)

\bibitem{youwang2023paintit}
Youwang, K., Oh, T.H., Pons-Moll, G.: Paint-it: Text-to-texture synthesis via deep convolutional texture map optimization and physically-based rendering. In: IEEE Conference on Computer Vision and Pattern Recognition (CVPR) (June 2024)

\bibitem{yu2023texture}
Yu, X., Dai, P., Li, W., Ma, L., Liu, Z., Qi, X.: Texture generation on 3d meshes with point-uv diffusion. In: Proceedings of the IEEE/CVF International Conference on Computer Vision. pp. 4206--4216 (2023)

\bibitem{ControlNet}
Zhang, L., Rao, A., Agrawala, M.: Adding conditional control to text-to-image diffusion models. In: Proceedings of the IEEE/CVF International Conference on Computer Vision. pp. 3836--3847 (2023)

\end{thebibliography}
